\documentclass[sigconf]{acmart}

\AtBeginDocument{%
  }

\copyrightyear{2024} 
\acmYear{2024} 
\setcopyright{rightsretained} 
\acmConference[HRI '24 Companion]{Companion of the 2024 ACM/IEEE
International Conference on Human-Robot Interaction}{March 11--14,
2024}{Boulder, CO, USA}
\acmBooktitle{Companion of the 2024 ACM/IEEE International Conference on
Human-Robot Interaction (HRI '24 Companion), March 11--14, 2024, Boulder,
CO, USA}
\acmDOI{10.1145/3610978.3640580}
\acmISBN{979-8-4007-0323-2/24/03}

\usepackage[latin1]{inputenc}           
\usepackage[list=true]{subcaption}
\usepackage{listings}
\usepackage{multirow}
\usepackage{cleveref}[2012/02/15]

\usepackage{ifthen}
\usepackage{tikz}
\usepackage{atbegshi,picture}
\newcommand{\myleftstd}{1in}

\newcommand\headertext{%
  \footnotesize Di Fu, Fares Abawi, Philipp Allgeuer, and Stefan Wermter. 2024. Human
Impression of Humanoid Robots Mirroring Social Cues. In Companion of
the 2024 ACM/IEEE International Conference on Human-Robot Interaction
(HRI '24 Companion), March 11-14, 2024, Boulder, CO, USA. \url{https://doi.org/10.1145/3610978.3640580}.
}

\newcommand{\setheader}{%
  \ifthenelse{\isodd{\thepage}}%
    {\newcommand{\myleftmargin}{\oddsidemargin+\myleftstd}}%
    {\newcommand{\myleftmargin}{\evensidemargin+\myleftstd}}%
    {{\oddsidemargin+\myleftstd}}%
    {{\evensidemargin+\myleftstd}}%
  \AtBeginShipoutNext{\AtBeginShipoutUpperLeft{%
    \put(\dimexpr\myleftmargin\relax,+1.2cm){\parbox{\textwidth}{\normalsize\sffamily\noindent\centering\hfill}}%
    \begin{tikzpicture}[remember picture,overlay]
      \node[anchor=north,yshift=-10pt] at (current page.north) {\parbox{\dimexpr\textwidth-\fboxsep-\fboxrule\relax}{\headertext}};
    \end{tikzpicture}%
  }}%
}
\let\citep\cite
\let\citet\cite
\let\citeyear\cite

\lstMakeShortInline[columns=fixed]|
\makeatletter
\gdef\@copyrightpermission{
  \begin{minipage}{0.3\columnwidth}
   \href{https://creativecommons.org/licenses/by/4.0/}{\includegraphics[width=0.90\textwidth]{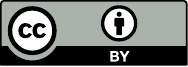}}
  \end{minipage}\hfill
  \begin{minipage}{0.7\columnwidth}
   \href{https://creativecommons.org/licenses/by/4.0/}{This work is licensed under a Creative Commons Attribution International 4.0 License.}
  \end{minipage}
  \vspace{5pt}
}
\makeatother
\begin{document}

\setheader

\title{Human Impression of Humanoid Robots Mirroring Social Cues}

%\title{Assessing the Perceived Impression of Social Robots Through Movement and Affective Mirroring}


\author{Di Fu}
\authornotemark[1]
\email{di.fu@uni-hamburg.de}
\affiliation{%
  \institution{University of Hamburg}
  \streetaddress{Vogt-Koelln-Str. 30}
  \city{Hamburg}
  \country{Germany}
}

\author{Fares Abawi}
\authornote{Both authors contributed equally to this research.}
\email{fares.abawi@uni-hamburg.de}
\affiliation{%
  \institution{University of Hamburg}
  \streetaddress{Vogt-Koelln-Str. 30}
  \city{Hamburg}
  \country{Germany}
}

\author{Philipp Allgeuer}
\email{philipp.allgeuer@uni-hamburg.de}
\affiliation{%
  \institution{University of Hamburg}
  \streetaddress{Vogt-Koelln-Str. 30}
  \city{Hamburg}
  \country{Germany}
}

\author{Stefan Wermter}
\email{stefan.wermter@uni-hamburg.de}
\affiliation{%
  \institution{University of Hamburg}
  \streetaddress{Vogt-Koelln-Str. 30}
  \city{Hamburg}
  \country{Germany}
}

% \renewcommand{\shortauthors}{Abawi et al.}

\begin{abstract}
Mirroring non-verbal social cues such as affect or movement can enhance human-human and human-robot interactions in the real world. The robotic platforms and control methods also impact people's perception of human-robot interaction. However, limited studies have compared robot imitation across different platforms and control methods. Our research addresses this gap by conducting two experiments comparing people's perception of affective mirroring between the iCub and Pepper robots and movement mirroring between vision-based iCub control and Inertial Measurement Unit (IMU)-based iCub control. We discovered that the iCub robot was perceived as more humanlike than the Pepper robot when mirroring affect. A vision-based controlled iCub outperformed the IMU-based controlled one in the movement mirroring task. Our findings suggest that different robotic platforms impact people's perception of robots' mirroring during  HRI. The control method also contributes to the robot's mirroring performance. Our work sheds light on the design and application of different humanoid robots in the real world.
\end{abstract}




\begin{teaserfigure}
  \centering
  \includegraphics[width=0.90\textwidth]{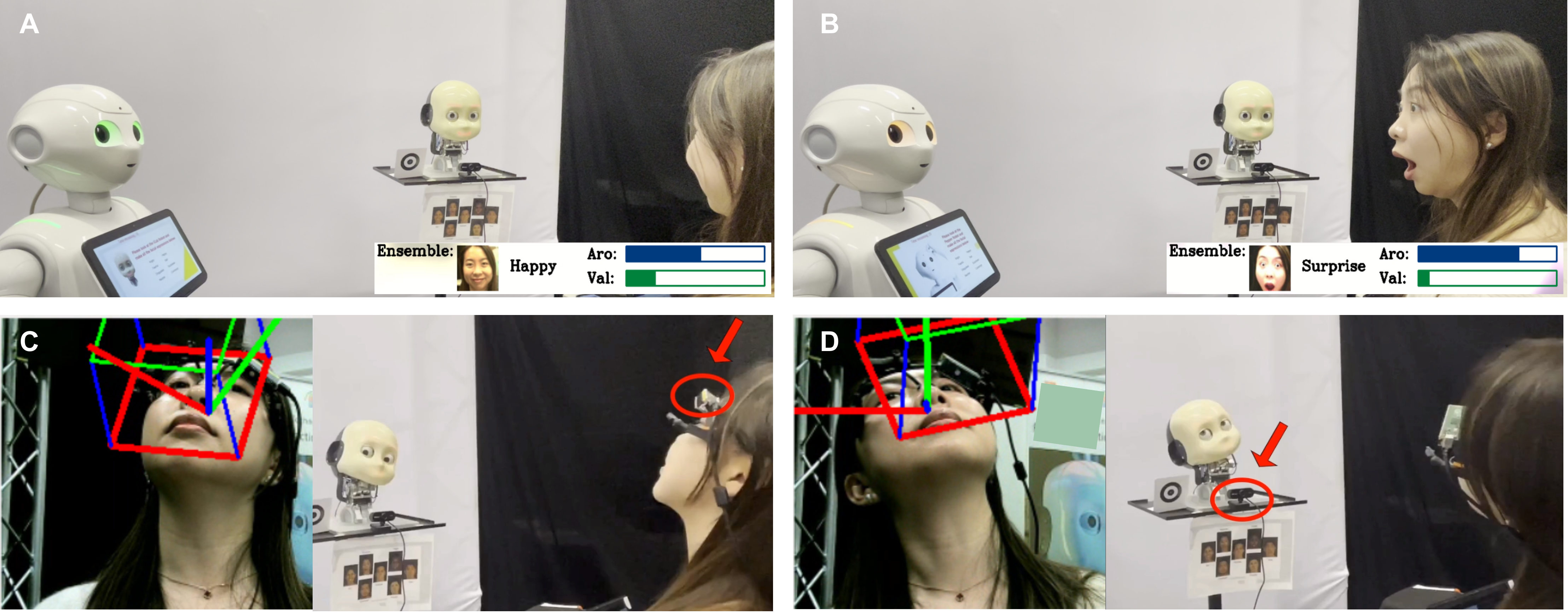}
  \vspace{-0.9em}
  \caption{A participant performing the four mirroring tasks in random order: A) The iCub robot mirroring facial expressions; B) The Pepper robot affectively signaling through LED color changes; C) The iCub robot mirroring head movement based on an inertial measurement unit (IMU) readings. The red circle shows the IMU; D) The iCub robot mirroring head movement according to a vision-based model. The red circle shows the camera.}
  \Description{A participant performing the four mirroring tasks in random order: A) The iCub robot mirroring facial expressions; B) The Pepper robot affectively signaling through LED color changes; C) The iCub robot mirroring head movement based on an Inertial Measurement Unit (IMU) readings. The red circle shows the IMU; D) The iCub robot mirroring head movement according to a vision-based model.The red circle shows the camera.}
  \label{fig:intro_teaser}
\end{teaserfigure}


\maketitle
%%%%%%%%%%%%%%%%%%%%%%%%%%%%%%%%%%%%%%%%%%%%%%%%%%%%%%%%%%%%%%%%%%%%%%%%%%%%%%%%

\section{Introduction}

The mirror neuron system (MNS) in humans facilitates the understanding of others by simulating their behaviors via sensorimotor processes~\citep{carr2014mirroring}. Mirroring, a fundamental element of social interaction, involves subconsciously imitating another individual's nonverbal cues, such as gestures, expressions, and postures~\citep{gergely2018social}. It can reflect an adaptive integration and utilization of social cues within the social context~\citep{van2003effects}. This mechanism often leads individuals to collaborate with those who exhibit similar and familiar behaviors~\citep{endedijk2017neural}. Mirror system dysfunction contributes to difficulties in social communication for individuals with Autism Spectrum Disorders (ASD)~\citep{prinsen2022broken}.  Mirroring also plays a significant role in human-robot social interaction. By mimicking non-verbal social cues, humans feel socially closer to the robot and perceive it as more aware of the intentions behind their social behaviors~\citep{li2015observer}.

% \begin{figure}[!htbp]
%         \centering
%         \includegraphics[width=0.46\textwidth]{src/imgs/participant_emotions_small.png}
%         \vspace*{-2mm}
%         \caption{The iCub robot mirroring facial expressions.}
%         \label{fig:experimental_setup}
%     \end{figure}
    
For robots, affective mirroring causes people to perceive the robot as an agent capable of conveying internal states, displaying social intelligence, and expressing humanlike characteristics~\citep{damiano2015towards,breazeal2016social}. Gonsior et al.~\citeyear{gonsior2011empathy} investigated the impact of mirroring facial expressions on empathy and perceived subjective performance in interactions with the robot head EDDIE~\citep{sosnowski2006eddie}, revealing that adaptive modes of robot behavior, where the robot mirrored human expressions, led to increased levels of human empathy and improved perceived task performance compared to a non-adaptive mode---without facial expression mimicry. Although most previous research shows consistent findings, few studies compare people's perceptions of affective mirroring on different humanoid robots. Robots convey emotional signals in various ways. For instance, the iCub robot can display simplified facial expressions with LED light
pattern changes, and the Pepper robot can change the color of the shoulder and eyelids to represent emotions. It may cause people to interpret them differently for the same expression. 

Movement mirroring enhances robots' sociability during human-robot interactions, making them more humanlike, empathetic, and socially intelligent~\citep{bugnariu2013human}. Two primary methods of enabling robots to mirror human movements include IMU-based controlled and vision-based controlled imitations. IMU-based controlled mirroring uses readings from an IMU attached to a head-mounted eye tracker worn by an actor to directly translate their head movements into robotic actions~\citep{geminiani2019design}. In contrast, vision-based controlled mirroring uses external cameras and pose estimation algorithms to interpret an actor's head movements and mirror them through a robot~\citep{ferro2019vision}. Liu et al.~\citeyear{liu2022real} show that the lightweight model surpasses the other state-of-the-art models on the same robot doing the head movement mirroring. Geminiani et al.~\citeyear{geminiani2019design} find that the Microsoft Kinect-based controlled NAO robot outperforms the IMU-based controlled NAO robot regarding limb movement mirroring in the autism treatment. However, comparing different control methods of robots on doing head and gaze mirroring remains to be studied.

%Riek et al.~\citeyear{riek2010robotsmiles} explored the effect of real-time head gesture mimicry on human-robot rapport, finding that full head gesture mimicking resulted in more positive interaction ratings compared to partial mimicking and non-mimicking. 

%Performance of robots' imitation behaviors is measured from multiple dimensions during human-robot interaction. For instance, some studies reported that affective imitation makes the robot more socially intelligent, humanlike, and less mechanical~\citep{buss2011towards}. This observation stems from the robots' capabilities to learn and adapt by understanding and responding to human emotions and social signals~\citep{kerzel2022s}. Metrics such as humanlikeness and responsiveness are used to assess how well robots' emotional reactions align with human anticipations. In the context of movement imitation, the mechanical and responsiveness criteria rate the precision, fluidity, and adaptability of robots as perceived by humans~\citep{fuente2015influence}. In this study, we evaluated robot performance based on four impression dimensions: social intelligence, mechanical attributes, responsiveness, and human likeness, as referenced in Seifert et al.~\citep{seifert2022imitating}. Moreover, we aim to explore the interconnections between these dimensions in various robot imitation tasks to gain deeper insights into their collective impact on human-robot interaction.

 \textcolor{black}{Social robots are designed to aid people, but individuals have been adapting to the robots instead. This is due to the fact that robots are not always designed with human preferences and interactive needs~\citep{lim2021social,vsabanovic2010robots}.  Researchers in robotic mirroring are constantly improving humanoid robots' accuracy and timeliness in simulating social cues. However, research about subjective evaluation and preference of the robotic platform and control method is limited.   } 

In this study, we conducted two experiments with two humanoids, the iCub and Pepper robots, as shown in ~\autoref{fig:intro_teaser}. The first experiment compared people's perceptions of affective mirroring on different humanoid robots. The second experiment assessed the impact of various control methods on the same robot platform doing movement mirroring. We evaluated the robots' performance by their mirroring speed and accuracy. People's perception of the robots was measured from four dimensions---Socially Intelligent, Mechanical, Responsive, and Humanlike. Through these investigations, our goal is to enhance the alignment of robotic design with human interaction preferences. We aim to solve these issues by investigating the following research questions (RQ): 

\begin{list}{RQ\arabic{enumi}}{\usecounter{enumi}\setlength{\leftmargin}{22px}}
    \item How do different robotics platforms, specifically the iCub and Pepper robots, compare in affective mirroring? 
    \item How do various robotic control methods, especially vision-based controlled and IMU-based controlled methods, impact the iCub robot's performance in movement mirroring tasks? 
\end{list}

%\section{Related Work}
% fabawi: Here we talk about studies that focus on the design, modeling and mimicry architecture in robotics
%Silva et al.~\citeyear{silva2016mirroring} devised a system enabling the Zeno RoboKind~\citep{hanson2009zeno} humanoid robot to replicate and mirror a user's facial expressions and head movements, captured using a 3D sensor in real-time. The system displayed high accuracy and resemblance in the synthesized expressions during interactions with typically developing children. Ondras et al.~\citeyear{ondras2017automatic} enabled the NAO~\citep{gouaillier2009nao} humanoid robot to replicate a teleoperator's facial expressions through affective signaling using LEDs around its eyes due to its static face, and head movements in real-time, with results showing successful communication of emotions and head movements, evidenced by high agreement among external observers.

% fabawi: Here we talk about studies that examine the effect of robot mimicry on humans

\section{Study Design}

\subsection{Affective Mirroring Task}
In this experiment, participants were asked to make eight facial expressions---\emph{Anger}, \emph{Fear}, \emph{Happiness}, \emph{Disgust}, \emph{Sadness}, \emph{Neutral}, \emph{Surprise}, and \emph{Contempt}---in front of the Pepper or iCub robots. The expressions were to be performed within one minute in any order. The robot mirrored participants' expressions either through \textit{affective signaling}---by changing the Pepper robot's eye and shoulder LED colors~\cite{lin2023experimental,johnson2013imitating}---or \textit{robotic facial expressions}---by changing the iCub robot's eyebrow and mouth LED patterns~\cite{aoki2022novel}. Next, participants were asked to match the colors displayed on the Pepper robot (depicted in the top row 
 of ~\autoref{fig:robot_emotions}) and facial expressions on the iCub robot (depicted in the bottom row 
 of ~\autoref{fig:robot_emotions}) to emotion categories. Technical details for running the experiment are provided as part of the Wrapyfi~\cite{abawi2024wrapyfi} tutorial series\footnote{\url{https://wrapyfi.readthedocs.io/en/latest/tutorials/Multiple\%20Robots.html}}.

Upon completion of the task, participants were asked to scan a QR code appearing on the Pepper's tablet using their cell phones to complete a three-item questionnaire, evaluating their experiences with either robot. In both questionnaires, participants were asked to rate their interaction with the robots using a 5-point Likert scale:
\begin{list}{Q\arabic{enumi}}{\usecounter{enumi}\setlength{\leftmargin}{17px}}
    \item How precise was the robot in mirroring your facial expressions? (1 = very imprecise, 5 = very precise)
    \item Did the robot mirror your expressions with major delay? (1 = no significant delay, 5 = significant delay)
\end{list}
\noindent Participants rated their impression of the robots on four dimensions ---\emph{Socially Intelligent}, \emph{Mechanical}, \emph{Responsive}, and \emph{Humanlike}---using a 5-point Likert scale (1 = not at all, 5 = yes, a lot).

\subsection{Gaze and Head Movement Mirroring Task}

In this experiment, participants interacted with the iCub robot given two conditions. Under the vision-based controlled condition, the iCub robot's movements were actuated by a vision-based head pose estimation model. Under the inertial measurement unit (IMU) controlled condition, the orientation readings arrived instead from an IMU attached to a wearable eye tracker. Participants wore the eye tracker and were asked to look at the iCub robot, freely moving their eyes and head. Participants observed the movements of the iCub robot to evaluate the interaction. Technical details for running the experiment are provided as part of the Wrapyfi~\citep{abawi2024wrapyfi} tutorial series\footnote{\url{https://wrapyfi.readthedocs.io/en/latest/tutorials/Multiple\%20Sensors.html}}. % While participants moved their heads, they had to ensure that the iCub robot was always within their field of view. 

Participants were asked to rate their interaction with the iCub robot using a 5-point Likert scale:

\begin{list}{Q\arabic{enumi}}{\usecounter{enumi}\setlength{\leftmargin}{17px}}
    \item How precise was the robot in mirroring your head movements? (1 = very imprecise, 5 = very precise)
    \item[Q2] Did the robot mirror your head movements with major delay? (1 = no
significant delay, 5 = significant delay)
    \item[Q3] Did the robot move its eyes? (Yes/No)
    \item[Q4] How precise was the robot in mirroring your eye movements? (1 = very imprecise, 5 = very precise)
    \item[Q5] Did the robot mirror your eye movements with major delay? (1 = no
significant delay, 5 = significant delay)
\end{list}

\noindent Participants rated their impression
of the iCub robot on four dimen\-sions---\emph{Socially Intelligent}, \emph{Mechanical},
\emph{Responsive}, and \emph{Humanlike}---using a 5-point Likert scale (1
= not at all, 5 = yes, a lot).

\subsection{Experimental Setup}
The participants were seated 80~cm away from the iCub robot's head, adjusting its height to match their eye level. A circular marker was placed beside the iCub robot to calibrate the Pupil Core eye tracker. Situated in front of the iCub robot was a Logitech C920 webcam facing the participants to perform tasks requiring a fixed view of their faces while the iCub robot moved its head and eyes. The Pepper robot stood facing the participants at an angle of 45 degrees with a distance of 1.2~m. The Pepper robot displayed an illustration of the ongoing task on its tablet and communicated the instructions verbally. The interaction was one minute long per task condition and the condition order was randomized. We used the Wrapyfi~\citep{abawi2024wrapyfi} framework for managing the task order, transmitting data between models and robots using various middleware, and orchestrating the experimental pipeline.

\subsection{Participants}
\textcolor{black}{30 participants (female = 7, male = 22, preferred not to say = 1) took part in both studies. Participants were between 24 and 41 years of age, with a mean age of 28.7. %Among the participants, 7 were Asian, 1 was Black, 3 were Middle Eastern or North African, 17 were White, and 2 reported other ethnicities. 
All participants reported no history of neurological conditions---seizures, epilepsy, stroke, etc.---and had normal or corrected-to-normal vision and hearing. One participant's data was excluded from the Pepper robot's affective mirroring experiment because of self-reported color blindness. Another participant's data was excluded from the iCub robot's movement mirroring experiment due to technical issues. This study adhered to the principles expressed in the Declaration of Helsinki. Participants signed consent forms approved by the Ethics Committee at the Department of Informatics, University of Hamburg.}

\begin{figure*}[!hbtp]
\centering\includegraphics[width=0.76\textwidth]{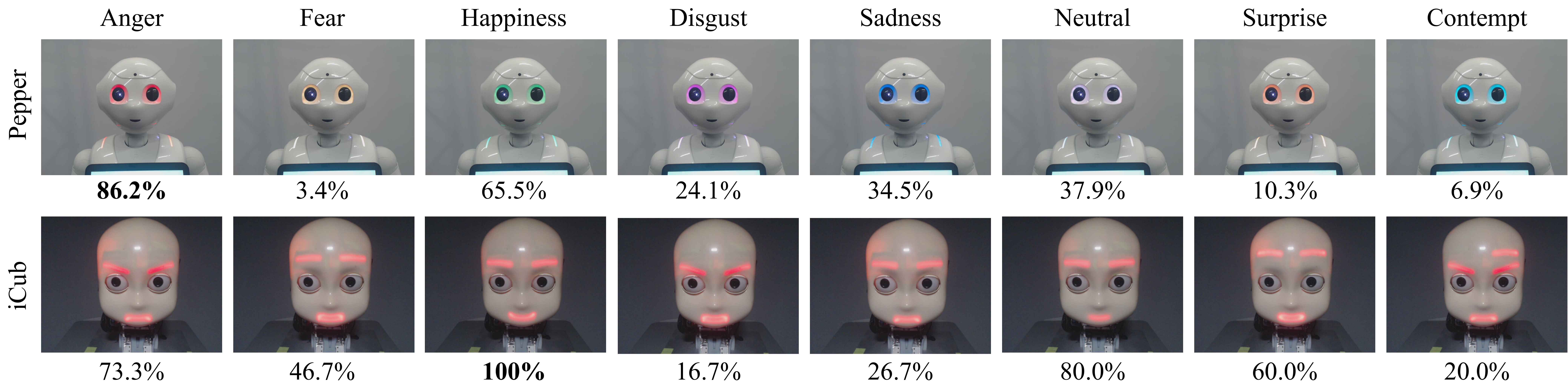}
\vspace*{-2mm}
  \caption{Eight emotion categories mimicked on the Pepper (Top) and iCub (Bottom) robots in the form of affective signaling and robotic facial expressions, respectively. Results of the human study are reported below each image in terms of the average accuracy in matching each affective signal or facial expression to an emotion category.}\label{fig:robot_emotions}
\end{figure*}

\section{Results}
We evaluated the results of both mirroring tasks, studying the perceived impression of the robot in each separate condition, as well as comparing the paired conditions within each respective task. Normality tests were conducted on the participants' answers to each dimension of the questionnaires. Results showed that their responses were normally distributed. In addition, all Post hoc tests in this study used Bonferroni correction.

\subsection{Affective Mirroring}

\textcolor{black}{For the affective mirroring task on either robot, the recognition accuracy is listed in ~\autoref{fig:robot_emotions}. For the Pepper robot, participants were most accurate in recognizing anger~(86.2\%) and least accurate in recognizing fear~(3.4\%). For the iCub robot, participants were most accurate in recognizing happiness (100\%) and least accurate in recognizing disgust~(16.7\%).}

\textcolor{black}{For participants' rating of interaction with the robots, results of paired-samples $t$-tests displayed no significant difference in precision (Q1) between the Pepper ($\text{mean} \, \pm \, \text{SE} = 2.79 \pm .18$) and iCub ($\text{mean} \, \pm \, \text{SE} = 2.90 \pm .15$) robots, ($t \left( 28 \right) = .46$, $p = .65$). No significant difference in delay (Q2) was found between the Pepper ($\text{mean} \, \pm \, \text{SE} = 2.38 \pm .18$) and iCub ($\text{mean} \, \pm \, \text{SE} = 2.48 \pm .20$) robots, ($t \left( 28 \right) = .52$, $p = .61$). For participants' rating of the impression of the robots, results of paired-samples $t$-tests displayed that the iCub ($\text{mean} \, \pm \, \text{SE} = 2.86 \pm .20$) robot was rated significantly more humanlike than the Pepper ($\text{mean} \, \pm \, \text{SE} = 2.10 \pm .16$) robot, ($t \left( 28 \right) = 3.45$, $p < .01$). No significant differences were found for the other three dimensions---\emph{Socially Intelligent}, \emph{Mechanical}, and \emph{Responsive}---between the two robots ($ps > .05$) (See ~\autoref{tab:impression_robots}). } 

\subsection{Movement Mirroring}
\textcolor{black}{A paired-samples $t$-tests showed that participants rated the vision-based controlled robot ($\text{mean} \, \pm \, \text{SE} = 3.55 \pm .24$) significantly more precise (Q1) than the IMU-based controlled robot ($\text{mean} \, \pm \, \text{SE} = 2.90 \pm .19$), ($t \left( 26 \right) = 2.19$, $p < .05$). The vision-based controlled robot ($\text{mean} \, \pm \, \text{SE} = 2.00 \pm .17$) was rated significantly less delayed (Q2) than the IMU-based controlled robot ($\text{mean} \, \pm \, \text{SE} = 2.66 \pm .21$), ($t \left( 26 \right) = -3.09$, $p < .01$). Under the vision-based controlled condition, all participants observed that the robot mirrored their eye movements, whereas two did not under the IMU-based controlled condition (Q3). Therefore, we only analyzed data from 27 participants who reported observing eye movement under both conditions. The paired-samples $t$-test showed no significant difference in the precision rating of the eye movement between the vision-based controlled robot ($\text{mean} \, \pm \, \text{SE} = 2.48 \pm .19$) and the IMU-based controlled robot ($\text{mean} \, \pm \, \text{SE} = 2.37 \pm .19$) ($p > .05$) (Q4). Also, no significant difference was found in the delay rating of the eye movement between the vision-based controlled robot ($\text{mean} \, \pm \, \text{SE} = 3.07 \pm .23$) and the IMU-based controlled robot ($\text{mean} \, \pm \, \text{SE} = 3.48 \pm .24$) ($p > .05$) (Q5). For the impression of the robot, participants reported that the vision-based controlled iCub ($\text{mean} \, \pm \, \text{SE} = 3.66 \pm .22$) robot was significantly more responsive than the IMU-controlled robot ($\text{mean} \, \pm \, \text{SE} = 3.17 \pm .21$), ($t \left( 26 \right) = 2.39$, $p < .05$). However, no significant differences were found in the remaining dimensions ---\emph{Socially Intelligent}, \emph{Mechanical}, and \emph{Humanlike}---between the two conditions ($ps > .05$) (results are shown in ~\autoref{tab:impression_robots}). } 

% fabawi: checked
%\subsection{Comparing Affective and Movement Mirroring}

%\begin{figure}[!hbtp]
%  \centering
  %\includegraphics[width=0.475\textwidth, trim = 9em 0 12em 0, clip=true]{src/imgs/plots_analyses/Correlation_matrix.pdf}
  %\vspace*{-2mm}
  %\caption{Correlation matrix illustrating the coefficients and significance between participant impressions of the robot under the two conditions in the mirroring tasks. \emph{M\_} refers to movement mirroring and \emph{E\_} to affective mirroring.}
  %\label{fig:correlation_conditions}
%\end{figure}

%\begin{figure*}[!hbtp]
%\centering
%\vfill
%FindWho
%\hfill
%\subcaptionbox{Socially Intelligent}
%{\includegraphics[width=0.323\textwidth, trim = 0.25em 0.31em 17.25em 3.65em, clip=true]{src/imgs/plots_analyses/Social intelligence.pdf}}%
%\hfill
%\subcaptionbox{Mechanical}{\includegraphics[width=0.22\textwidth, trim = 12em 0 17.25em 5em, clip=true]{src/imgs/plots_analyses/Mechanical.pdf}}%
%\hfill
%\subcaptionbox{Responsive}{\includegraphics[width=0.22\textwidth, trim = 12em 0 17.25em 5em, clip=true]{src/imgs/plots_analyses/Responsive.pdf}}%
%\hfill
%\subcaptionbox{Humanlike}{\includegraphics[width=0.218\textwidth, trim= 12em 0 17.25em 5em, clip=true]{src/imgs/plots_analyses/Humanlike.pdf}}%
%\caption{Participants' impressions of robots in different affective and movement mirroring conditions. \\ $*$ \textmd{denotes} $.01 < p < .05$, $*\!*$ $.001 < p < .01$, $*\!*\!*$ $p < .001$, \textmd{and \textit{n.s.} denotes no significance.}}
%\end{figure*}

\begin{figure*}[!hbtp]
\centering
\vfill
\subcaptionbox{Affective Mirroring}{\includegraphics[width=0.45\textwidth, trim= 0 0 0 2.3em, clip=true]{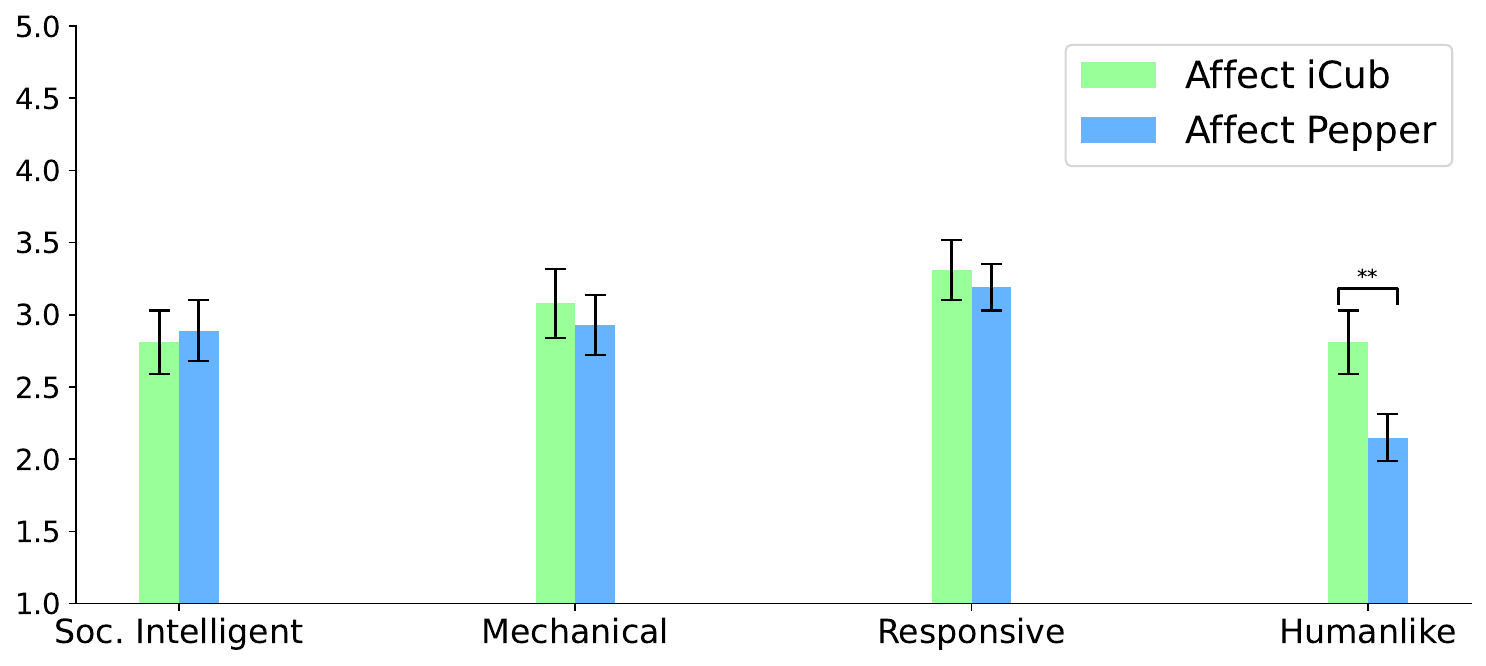}}%
\hfill
\subcaptionbox{Movement Mirroring}{\includegraphics[width=0.45\textwidth, trim = 0 0 0 2.3em, clip=true]{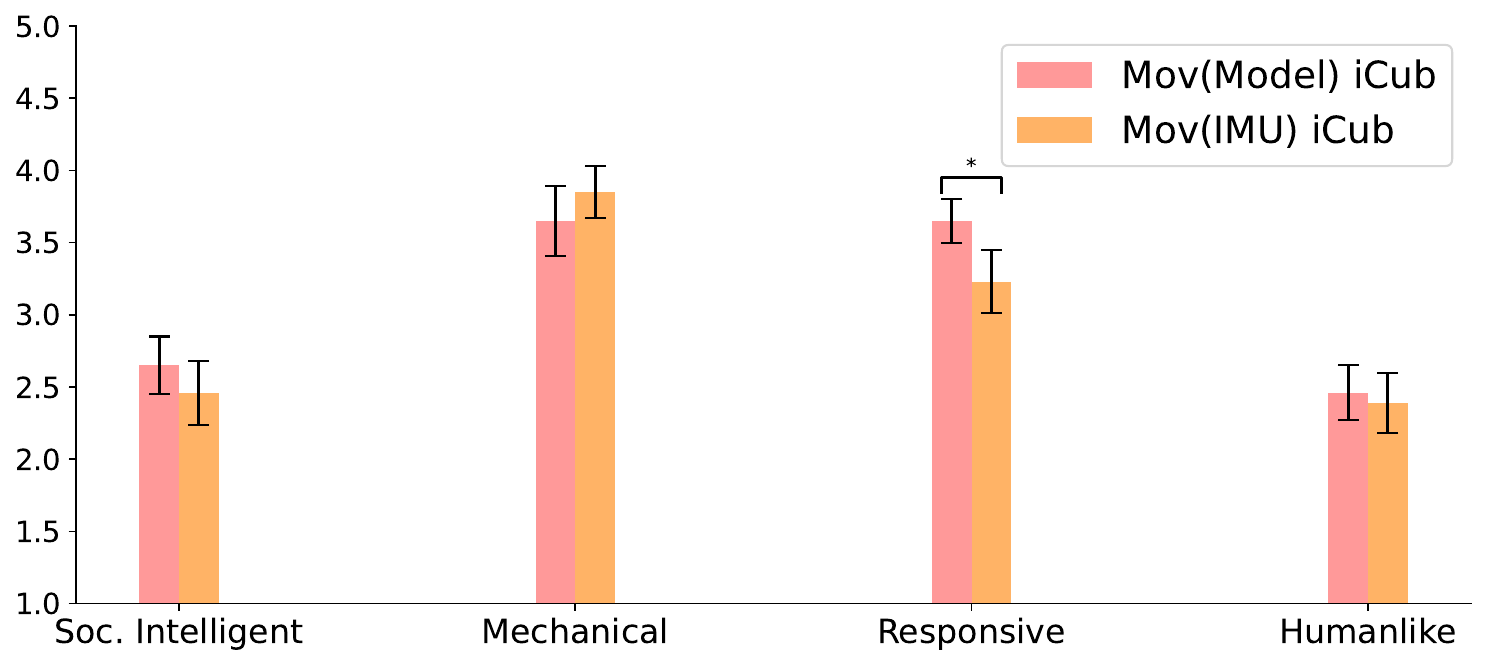}}%
\caption{Participants' impressions (5-point Likert scale) of robots under different affective and movement mirroring conditions. \\ \footnotesize{$*$ \textmd{denotes} $.01 < p < .05$\text{, and} $*\!*$ $.001 < p < .01$}}
\end{figure*}

\begin{table}[!hbtp]
\centering
\caption{Impression of the robots under different task conditions (Mean $\pm$ SE)}
\label{tab:impression_robots}
\resizebox{\columnwidth}{!}{ % Resizing the table to fit within a single column width
\begin{tabular}{ccccc}
\hline
 & \multicolumn{1}{c}{Affect} & \multicolumn{1}{c}{Affect} & \multicolumn{1}{c}{Mov.(Model)} & \multicolumn{1}{c}{Mov. (IMU)} \\
 & \multicolumn{1}{c}{iCub} & \multicolumn{1}{c}{Pepper} & \multicolumn{1}{c}{iCub}  & \multicolumn{1}{c}{iCub}  \\ \hline

Soc. Intelligent    & 2.81 $\pm$ .22 & 2.89 $\pm$ .21 & 2.65 $\pm$ .20 & 2.46 $\pm$ .22 \\
Mechanical          & 3.08 $\pm$ .24 & 2.93 $\pm$ .21 & 3.65 $\pm$ .24 & 3.85 $\pm$ .18 \\
Responsive          & 3.31 $\pm$ .21 & 3.19 $\pm$ .16 & 3.65 $\pm$ .15 & 3.23 $\pm$ .22 \\
Humanlike           & 2.81 $\pm$ .22 & 2.15 $\pm$ .16 & 2.46 $\pm$ .19 & 2.39 $\pm$ .21 \\ \hline
\end{tabular}}
\end{table}

\section{Discussion}
\textcolor{black}{Participants associated the iCub robot's facial expressions with emotions more than the Pepper robot's affective signaling and found the iCub robot more humanlike. Another observation relates to the accuracy of recognizing different affective signals conveyed by either robot. Participants could accurately associate \emph{Anger} with the color red and \emph{Happiness} with green on the Pepper robot. This is complemented by findings associating exposure to different colors with physiological and psychological responses~\citep{wilms2018color,song2017expressing}. Participants more accurately identified expressions of \emph{Happiness}, \emph{Neutral}, and \emph{Surprise} on the iCub robot compared to the Pepper robot. This can be attributed to humans primarily relying on observing the mouth and eyebrows to recognize these facial expressions~\citep{guarnera2017facial}, features that the Pepper robot lacks.}

We compared two movement mirroring methods. The vision-based controlled method produced smoother, more precise, and more responsive movements than the IMU-based controlled method. The IMU-based 
controlled method transfers the IMU readings at a faster rate,
but this causes jittery movements due to hardware limitations. These findings are also consistent with Geminiai et al.~\citep{geminiani2019design} that the IMU-based NAO robot is more intrusive and requires longer setup time than the Kinect-based NAO robot during the limb movement mirroring. However, in our study, both methods were perceived as equally humanlike, implying that less responsiveness does not contradict humanlikeness. 

%\textcolor{black}{Comparing affective and movement mirroring tasks showed that the robots in the affective mirroring task were more socially intelligent and less mechanical than the ones in the movement mirroring task. These interesting results might imply that for human-robot social interaction, affective mirroring might be valued more significantly compared with movement mirroring. The findings can be used to inspire future robot designs. To build an intelligent robot, they should be endowed with the ability to recognize and mirror human affective behaviors. No significant differences between responsiveness and humanlikeness were found between the tasks. }

%\textcolor{black}{In both tasks, participants' perceptions of robot social intelligence are positively related to their responsive and humanlike levels. However, the robot's responsive and humanlike levels were only positively correlated in the affective mirroring task and not in the movement mirroring task. These results implied that during movement mirroring, the robot is not perceived in the sense that the faster it reacts, the more humanlike it is. However, in the affective mirroring task, \emph{Responsive} is still a key factor related to \emph{Humanlike} for the robot. This may be because affective imitation is expected to be quick and reflexive, whether in human-human interaction or human-agent interaction 
% \cite{hrdy2020emergence}.}
 
 \textcolor{black}{Several limitations could be addressed and investigated in future research. We could not compare movement mirroring on the two humanoid robots. This is because the Pepper robot is not able to roll its head or move its eyes, unlike the iCub robot. Our iCub robot doesn't have a full body, hence, we cannot study the limb mirroring between the two robots. Future studies could address the interaction effect between affective and movement mirroring. Moreover, researchers could investigate how different humanoid robots and control methods impact children with ASD, and whether it affects their social functions~\citep{zheng2015robot}.}
%Second, more perceived impression dimensions of the robots could be tested, such as \emph{Likable} and \emph{Empathetic}, to help researchers better understand users' multi-dimensional preferences. Third, the complexity of the tasks could increase

\section{Conclusions}

We investigated human perceptions of two humanoid robots in the affective and movement mirroring tasks. Our findings revealed that a robot displaying facial expressions like an iCub robot was perceived as more humanlike than a robot conveying affective signals like a Pepper robot. For gaze and head mirroring, a vision-based controlled robot performed better than an IMU-based controlled robot. This could be attributed to latency in processing and transmitting the filtered IMU readings. In summary, we showed that robotic platforms and robot control methods played an essential role in mirroring tasks during HRI. It may guide future humanoid robot design decisions to align with humans' needs.

\begin{acks}
The authors gratefully acknowledge partial support from the German Research Foundation DFG under project CML~(TRR~169).
\end{acks}

\balance
\bibliographystyle{ACM-Reference-Format}
\bibliography{submission}

% \appendix









\end{document}